\documentclass[lettersize,journal]{IEEEtran}
\usepackage{amsmath,amsfonts}
\usepackage{algorithmic}
\usepackage{array}
\usepackage[caption=false,font=normalsize,labelfont=sf,textfont=sf]{subfig}
\usepackage{textcomp}
\usepackage{stfloats}
\usepackage{url}
\usepackage{verbatim}
\usepackage{graphicx}
\usepackage{balance}

\usepackage{bbm}
\usepackage{booktabs}
\usepackage{threeparttable}
\usepackage{multirow}
\usepackage{tabu}
\usepackage{tikz}
\usepackage{comment}
\usepackage{amsmath,amssymb} 
\usepackage{color}
\usepackage{cite}
\usepackage[linesnumbered,boxed,ruled,commentsnumbered]{algorithm2e}
\usepackage{algorithmic}

\usepackage[accsupp]{axessibility}  

\def\BibTeX{{\rm B\kern-.05em{\sc i\kern-.025em b}\kern-.08em
    T\kern-.1667em\lower.7ex\hbox{E}\kern-.125emX}}
\usepackage{balance}
\begin{document}
\title{MetaComp: Learning to Adapt for Online Depth Completion}
\author{Yang Chen, Shanshan Zhao, Wei Ji, Mingming Gong, Liping Xie
\thanks{Yang Chen and Liping Xie are with Key Laboratory of Measurement and Control of Complex System of Engineering, Ministry of Education, School of Automation, Southeast University, Nanjing 210096, China (e-mail: chenyang2111@gmail.com; lpxie@seu.edu.cn). Liping Xie is the corresponding author.}
\thanks{Shanshan Zhao is with the JD Explore Academy, Beijing, China (e-mail: sshan.zhao00@gmail.com).}
\thanks{Wei Ji is with National University of Singapore, Singapore (e-mail: jiwei@nus.edu.sg).}
\thanks{Mingming Gong is with School of Mathematics and Statistics, University of Melbourne, Australia (e-mail: mingming.gong@unimelb.edu.au).}
\thanks{This work was done during Yang Chen’s internship at JD Explore Academy.}
}


\maketitle

\begin{abstract}
Relying on deep supervised or self-supervised learning, previous methods for depth completion from paired single image and sparse depth data have achieved impressive performance in recent years. However, facing a new environment where the test data occurs online and differs from the training data in the RGB image content and depth sparsity, the trained model might suffer severe performance drop.
To encourage the trained model to work well in such conditions, we expect it to be capable of adapting to the new environment continuously and effectively. 
To achieve this, we propose MetaComp. It utilizes the meta-learning technique to simulate adaptation policies during the training phase, and then adapts the model to new environments in a self-supervised manner in testing.
Considering that the input is multi-modal data, it would be challenging to adapt a model to variations in two modalities simultaneously, due to significant differences in structure and form of the two modal data.
Therefore, we further propose to disentangle the adaptation procedure in the basic meta-learning training into two steps, the first one focusing on the depth sparsity while the second attending to the image content. During testing, we take the same strategy to adapt the model online to new multi-modal data.
Experimental results and comprehensive ablations show that our MetaComp is capable of adapting to the depth completion in a new environment effectively and robust to changes in different modalities.
\end{abstract}


\section{Introduction}
\IEEEPARstart{D}{epth} information is fundamental in various 3D vision tasks, including 3D object detection\cite{xu2018pointfusion}, 3D reconstruction\cite{alexiadis2012real}, 3D mapping\cite{newcombe2011kinectfusion, zhang2014loam} and human pose estimation\cite{moon2018v2v, luo2019real}. 
The depth data can be obtained by exploiting depth sensors, \textit{e.g.,} Lidar. However, compared with the RGB image data, the captured depth data is usually sparse, while in some scenarios, dense depth data is required to provide more complete 3D information.
Therefore, depth completion from paired single RGB image and sparse depth map has drawn much attention in recent years~\cite{cspn++,qiu2019deeplidar,acmnet,penet}.
Relying on large amounts of training data with ground-truth dense depth map, deep models trained via supervised learning have achieved remarkable performance for depth completion~\cite{acmnet, penet}. To reduce reliance on ground-truth depth, some works~\cite{calib,void} explore self-supervised methods that exploit the photometric loss and sparse depth consistency loss. All of these existing methods focus on the performance improvement for depth completion when the test data and training data come from the same environment and all test data are available offline. In this paper, we investigate the self-supervised depth completion under a new setting, where the trained model is deployed in a new and dynamic environment. This is a practical scenario in real-world applications, like autonomous driving.

In this new setting, existing depth completion models 
tend to 
suffer performance drops
when exposed to a new environment. 
The degradation is fatal for many tasks that require precise depth maps. 
For example, an autonomous driving car might encounter different 
conditions, such as weather, lighting, and scene, 
which result in dynamic content\footnote{In fact, images in different domains often vary, such as in appearance (or style) and scene structure. Here, we use the term ``content'' to express them.} changes in images it captured.
In addition, 
when deployed in a new environment, due to the sensor differences, such as hardware and assembly position, the model will be required to process sparse depth data with different sparsity.
Moreover, the occlusion between objects can also lead to 
the variation of density and distribution of sparse depth points captured.
If the model can not quickly modify itself to adapt to changes of environment, incorrect depth values would be predicted, which might lead to serious accidents.


Online adaptation is a profitable method to mitigate the performance degradation.
Fine-tuning a pre-trained model online on a video can help the model adapt to a new environment and thus achieve a good performance. 
Recently, unsupervised online adaptation has been widely studied on stereo depth, monocular depth estimation, and visual odometry \cite{l2a, selfvo, kuznietsov2021comoda}.
In comparison with these tasks which only explore the single modality, depth completion introduces multi-modal inputs containing paired single RGB image and sparse LiDAR data, which raises new opportunities and challenges for model adaptation. One the one hand, 
the sparse depth information can provide an additional geometric supervision, which would benefit the online adaptation procedure.
On the other hand, the involvement of sparse depth data also requires the model to adapt to different sparsity as well as the changing image contents.
To enable the pre-trained model to be capable of fast adapting to a new environment, we expect it to possess such capability during the pre-training on an offline dataset. This can be achieved by exploiting the meta-learning technique~\cite{hospedales2021meta}, which encourages the model to learn as it performs. 

Motivated by the analysis, we propose MetaComp, which is built on Model Agnostic Meta-Learning (MAML) \cite{maml}, a classical ``learning to learn" algorithm.
In detail, we fuse the process of online adaptation into the training paradigm based on MAML to learn a base model suitable for online depth completion.
To ease the adaptation to multi-modality, we propose to disentangle the adaptation procedure to two steps, one focusing on the sparse data while the other attempting to adapt to the image.
As a result, MetaComp provides a disentangled meta-learning algorithm, which divides the meta-training stage in MAML into two steps.
In the first step, which focuses on the adaptation to the sparse depth data, we exploit a sparse-to-dense network to recover a coarse dense depth map from the sparse depth data. We meta-train this module to adapt to different depth sparsity. 
In the second step, we take the coarse dense depth map and image as input and then meta-train another depth estimation network and camera pose estimation network for variation of images. Since the depth data becomes dense, the second step would focus on the image content change. We show the effectiveness of MetaComp through providing comprehensive comparisons.

In summary, our contributions are two-fold: 1) To deal with the adaptation to the new dynamic environment for depth completion, we propose to exploit the meta-learning strategy to learn a base model and alleviate the degradation in performance in a new environment during testing by online tuning the base model;
2) Considering the multi-modality of inputs, we improve the basic meta-learning algorithm by developing a disentangled meta-learning strategy to train our model and further enhance both its capability of adapting to a new environment with multi-modal input and robustness to changes in different modalities.

\section{Related work}
In this section, we review previous works related to depth completion, online adaptation, and meta-learning, respectively.
\noindent\textbf{Supervised depth completion.} 
Depth completion is aimed to recover dense depth maps from sparse depth maps, with 
\cite{liu2013guided, qiu2019deeplidar} or without \cite{sparse1, uncertainty} the guidance of corresponding images.
In real world, sparse depth maps are acquired by the depth sensor, \textit{e.g.,} LiDAR, yet, the distribution and density of the sparse depth data are irregular, which brings obstacles to depth completion.
To address the challenges from sparsity, several sparse invariant convolutions \cite{sparse1, sparse2, sparse3, sparse4} are proposed to process the sparse data instead of using the standard convolution operation.
To take advantage of the known depth values, spatial propagation strategy has been studied widely\cite{cspn,cspn++,nlspn,lin2022dynamic}, which recovers depth by propagating the known depth information spatially.
In addition, some methods~\cite{van2019sparse,penet,acmnet,fcfrnet} attempt to propose more effective fusion strategy to combine the dense (RGB image) and sparse (sparse depth) data. For example, Gansbeke \textit{et al.}~\cite{van2019sparse} proposed two sub-networks with confidence learning, one learning guidance from image and depth data to guide the other. Hu \textit{et al.}~\cite{penet} followed a similar clue, but used the spatial propagation strategy to refine the estimated depth map.

\noindent\textbf{Self-supervised or unsupervised depth completion.} 
Self-supervised or unsupervised depth completion assumes that the ground-truth dense depth is not available.
To address this task, a main solution is minimizing the depth loss between the predicted depth and sparse map, and the photometric loss between the RBG image and its warped image \cite{ss2d, calib, defusenet, void, adafram}.
For example, Ma \textit{et al.} \cite{ss2d} utilized Perspective-n-Point (PnP) \cite{pnp} and Random Sample Consensus (RANSAC) \cite{ransac} to estimate the relative pose between two adjacent frames and warp image. However, PnP and RANSAC are sensitive to the number of sparse input points and might fail when the number is too small.
In comparison, Wong et al. \cite{void} introduced an additional pose estimation network to reconstruct images and designed a piece-wise
planar scaffolding of the scene to predict a coarse dense depth map from sparse input. 
Recently, Wong \textit{et al.}~\cite{calib} added the intrinsic calibration parameters of the camera to input to increase generalization to different cameras. However, the changes of scenes are not considered in this method.
In addition, there are some methods using other cues to achieve self-supervised learning.
For instance, Yang \textit{et al.} \cite{ddp} pre-trained a conditional prior network (CPN) in a supervised manner on an additional dataset to obtain a strong prior and then trained the main model on the target dataset without dense depth maps relying on CPN.
Lopez-Rodrigue \textit{et al.} \cite{proada} introduced a domain adaptation technique for depth completion that requires ground truth of the additional dataset to pre-train the model.
In fact, sparse depth data has been able to provide a strong prior for depth completion.
In contrast to the above methods, our model is designed to adapt quickly to other domains, relying only on an additional dataset with sparse depth maps and images for pre-training. 

\noindent\textbf{Online adaptation.} 
Deep models in machine learning easily suffer from performance degradation, when exposed to a new environment.
To alleviate this domain shift, online adaptation is employed to adapt the model to the new environment \cite{knowles2021toward, kuznietsov2021comoda, zhong2018open, l2a, luo2018real}. 
Tonioni \textit{et al.} \cite{unstereo} proposed an adaption strategy for deep stereo based on gradients in an unsupervised way.
Casser \textit{et al.} \cite{casser2019depth} introduced camera ego-motion and object motions to unsupervised monocular depth prediction and designed an online refinement method to adapt the network to unknown domain.
Chen \textit{et al.}~\cite{chen2019self} combined tasks of predicting depth, optical flow, camera pose, and intrinsic parameters to a comprehensive model (GLNet), and proposed two online refinement methods to optimize model parameters or the output.
MADNet \cite{madnet} realized real-time adaptation of the model by updating a subset of parameters selectively.
Li \textit{et al.} \cite{selfvo} adopted a new online adaptation method using meta-learning based on the context of frames in a video in self-supervised visual odometry and utilized an online feature alignment method for Batch Normalization (BN) layer.
We apply online adaptation to combat domain changes on depth completion, and the prior and supervision provided by sparse maps can make our adaptation more reliable.

\noindent\textbf{Meta-learning.} 
Meta-learning, also called ``learning to learn'', is proposed to learn rules or algorithms from the structure in data or other algorithms \cite{mishra2017simple, thrun1998learning, gama2022weakly, li2021attribute}. 
Recent works in meta-learning are focused on rapidly updating models to novel domains \cite{maml, reptile, reptile2}.
Finn \textit{et al.} \cite{maml} raised Model Agnostic Meta-Learning (MAML) that pre-trains a model through updating the network indirectly using latent gradients to make the model easy to fine-tune and generalize.
Due to the excellent performance of MAML, it has been widely applied to increase the capability of the model to generalize and adapt to different tasks \cite{l2a, forget, selfvo, nagabandi2018deep}. Tonioni \textit{et al.} \cite{l2a} incorporated unsupervised training and supervised training to meta-learning to learn a model suitable for adaptation in stereo depth. 
Zhang \textit{et al.} \cite{forget} designed an adapter module trained by meta-learning to prevent catastrophic forgetting in monocular videos. 
In contrast against their works only exploring one modality, our meta-learning algorithm is focused on self-supervised depth completion with multi-modal inputs, and aimed to enhance the adaptability of the model to multi-modal changes in a dynamic environment.

\section{Method}

In this section, we first introduce our self-supervised framework including the network structure and loss function employed in our method.
Then, we illustrate the basic meta-learning algorithm and our disentangled meta-learning algorithm.
Finally, we elaborate the process of online adaptation.

\subsection{Self-supervised framework}
We achieve self-supervised depth completion through training the network on a series of consecutive two frames, using a sparse map of the current frame and a pair of images of the two frames.
For the $t$ frame, our model is aimed to predict a dense depth map $P_t$ from sparse depth map $S_t$ and RGB image $I_t$. 
Supervised depth completion employs the discrepancy between the predicted depth map and ground truth as loss function to train the model directly.
For self-supervised depth completion, due to the lack of ground truth, models are learned implicitly through minimizing the discrepancy between the predicted depth map and sparse depth input as well as the discrepancy between the input image and reconstructed image.
The reconstructed image could be acquired through warping the image of the adjacent frame, \textit{e.g.,} $t-1$, to the current according to the relative pose ${T}_{t\rightarrow t-1}$, \textit{i.e.,} the camera pose change from the $t$ frame to the $t-1$ frame.
Our network consists of three parts, including:  1) CoarseNet aiming at predicting a coarse dense depth map from the sparse depth map, 2) DepthNet aimed to estimating the final dense depth map from the RGB image and coarse dense depth map, and 3) PoseNet for relative camera pose estimation. Below, we introduce the network architecture and self-supervised loss function. 

\begin{figure*}[!t]
    \centering
    \includegraphics[scale=0.45]{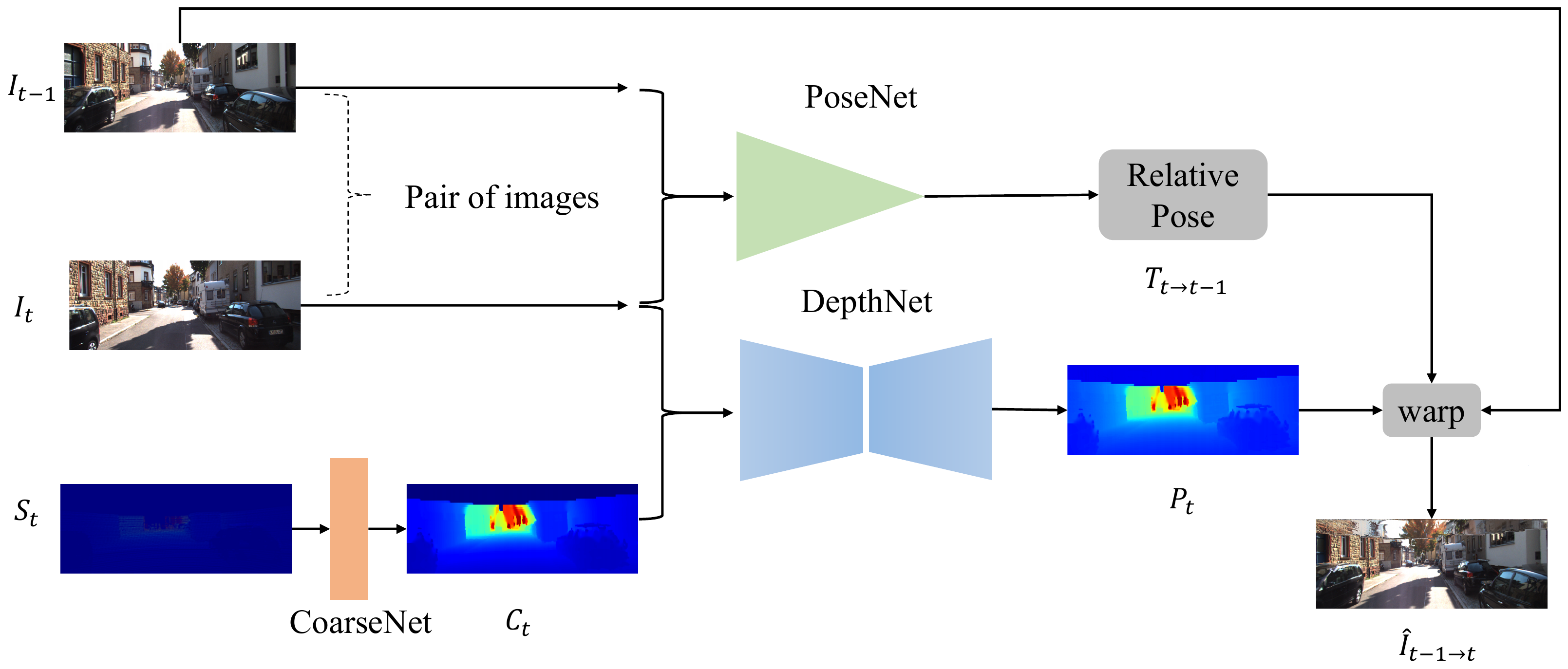}
    \caption{Network structure of self-supervised framework for depth completion. Our self-supervised framework consists of CoarseNet, PoseNet and DepthNet to predict coarse dense map, relative pose and final depth map, respectively. Warped image is generated by $P_t$, $I_{t-1}$ and $T_{t\rightarrow t-1}$.}
    \label{fig:net}
\end{figure*}

\noindent\textbf{Network structure.}
We use the network in \cite{ss2d} as our DepthNet to predict the final depth map.
Yet, it utilizes PnP and RANSAC algorithm to estimate the camera pose, which might fail when the number of sparse points is too small. 
Therefore, we adopt a pose estimation network with the structure used in \cite{dig}, \textit{i.e.,} PoseNet,  to estimate the relative pose of two adjacent frames from two images.
In addition, DepthNet directly utilizes traditional convolutions without considering the sparsity of depth input, which leads to the sensitivity to sparse inputs.
To enhance the robustness of our model to sparse input, we employ a series of normalized convolutions in \cite{ncnn} as our CoarseNet to predict an initial coarse dense map from a sparse map, which makes the depth input of DepthNet dense. 
The structure of our network is shown in Fig.~\ref{fig:net}.

\noindent\textbf{Loss function.}
Self-supervised loss function in depth completion usually consists of three parts, including a sparse depth consistency loss $\mathcal{L}_{sd}$, a photometric loss $\mathcal{L}_{ph}$, and a smoothness loss $\mathcal{L}_{sm}$ \cite{ss2d, calib, void}. 
The three losses and the whole loss are listed as follows:
\begin{equation}
\begin{aligned}
    &\mathcal{L}_{sd}(P_t, S_t) = ||\mathbbm{1}_{S_t>0}\cdot (P_t - S_t)||_2^2,\\
    & \mathcal{L}_{ph}(\hat{I}_{t}, I_{t}) = \frac{\alpha}{2}(1-SSIM(\hat{I}_{t}, I_{t}))+(1-\alpha)||\hat{I}_{t}-I_{t}||_1 ,\\
    & \mathcal{L}_{sm}(P_t) = ||\nabla^2P_t||_1,\\
    & \mathcal{L} = w_{sd}\mathcal{L}_{sd}(P_t, S_t) + w_{ph} \mathcal{L}_{ph}(\hat{I}_{t}(P_t, I_{t-1}, T_{t\rightarrow t-1}), I_{t}) \\
    & \qquad + w_{sm}\mathcal{L}_{sm}(P_t),
\end{aligned}
\end{equation}
where $\hat{I}_{t}(P_t, I_{t-1}, T_{t\rightarrow t-1})$ is the warped RGB image from  the $t-1$ frame to the $t$ frame\cite{dig}, SSIM is in \cite{wang2004image} with $\alpha$ set to $0.85$, and $w_{sd}$, $w_{ph}$, and $w_{sm}$ are all loss weights. 
As illustrated in Fig.~\ref{fig:net}, our model produces a coarse dense map $C_t$ first, then combines the image and coarse map as input to predict dense depth $P_t$, and estimates the relative pose $T_{t\rightarrow t-1}$ to warp image and compute losses.

\subsection{MetaComp}
To enable our model to adapt to a new environment online effectively, we train our model using the meta-learning algorithm.
In this part, we first introduce the basic meta-learning algorithm (BML) with two stages of meta-training and meta-testing to learn a base model suitable for self-supervised depth completion online adaptation. BML directly optimizes the whole model according to the input, which ignores the differences in the structure and form of two modal data.
To alleviate this issue, we propose disentangled meta-learning algorithm (DML) through disentangling the meta-training stage in BML into two steps, one adjusting the CoarseNet for sparse input while the other one adjusting the DepthNet and PoseNet for image input after the adaptation of CoarseNet.
In the following, we detail BML and DML, respectively.

\noindent\textbf{Basic meta-learning.}
To learn a model which is able to fast adapt itself to an unseen online video, we adjust the training mode by exploring a strategy similar to Model Agnostic Meta Learning (MAML) \cite{maml}. 
MAML is a recent popular meta-learning algorithm focused on learning to adapt to new tasks.
It divides the training process into two stages, including meta-training and meta-testing, which are also called inner-loop and outer-loop.
Let $\mathcal{D}$ be the training set including $T$ tasks and split $\mathcal{D}_{T}$ into $\mathcal{D}^{train}_{\tau \in T}$ and $\mathcal{D}^{test}_{\tau \in T}$ for each task $\tau$.
Thus, the objective of MAML can be expressed as follows:
\begin{equation}
    \underset{\theta}{\min} \sum_{\tau \in T} \mathcal{L}_{out}(\theta-\alpha\nabla_{\theta}\mathcal{L}_{in}(\theta, \mathcal{D}^{train}_{\tau \in T}), \mathcal{D}^{test}_{\tau \in T}),
\end{equation}
where $\mathcal{L}_{in}$ is the loss function used in inner-loop, $\mathcal{L}_{out}$ is the loss function used in outer-loop, $\theta$ is the parameter of the model and $\alpha$ is the learning rate for inner adaptation \cite{maml}.
Parameters of model are modified twice in each optimization in MAML.
In meta-training, MAML utilizes stochastic gradient descent (SGD) to adapt parameters for each task using $\mathcal{L}_{in}$.
Then, in meta-testing, it uses adapted parameters to test the model and optimize the original model based on testing performance.
The purpose of meta-training model on $\mathcal{D}^{train}$ is to make the model achieve a better performance on $\mathcal{D}^{test}$ in meta-testing.
Meanwhile, our online adaptation aims to adapt the model in the current frame and make it achieve a better performance in the next frame.
Therefore, we set our learning objective as follows:
\begin{equation}
\underset{\theta}{\min} \sum_{t} \mathcal{L}_{out}(\theta-\alpha\nabla_{\theta}\mathcal{L}_{in}(\theta, \mathcal{D}_{t}), \mathcal{D}_{t+1}),    
\end{equation}
where $\mathcal{D}_t$ contains a pair of images of frames $t-1$ and $t$ and a sparse depth map of frame $t$ in a video, namely, $\mathcal{D}_t=\{{I_{t-1}, I_{t}, S_{t}}\}$.
We adopt self-supervised loss of predicted final depth in Eq.~\ref{eq:lossf} as our \textit{outer loss}.
\begin{equation}
\begin{aligned}
    & \mathcal{L}_f = w_{sd}\mathcal{L}_{sd}(P_t, S_t) + w_{ph} \mathcal{L}_{ph}(\hat{I}_{t}(P_t, I_{t-1}, T_{t\rightarrow t-1}), I_{t})\\
    & \qquad + w_{sm}\mathcal{L}_{sm}(P_t).\\
\end{aligned}
\label{eq:lossf}
\end{equation}
To supervise the learning of CoarseNet, we also use
the loss with same formulation as Eq.~\ref{eq:lossf} to optimize it in meta-training:
\begin{equation}
\begin{aligned}
    & \mathcal{L}_c = w_{csd}\mathcal{L}_{sd}(C_t, S_t) + w_{cph} \mathcal{L}_{ph}(\hat{I}_{t}(C_t, I_{t-1},T_{t\rightarrow t-1}), I_{t}),\\
\end{aligned}
\label{eq:lossc}
\end{equation}
where $C_t$ is the predicted coarse dense map of CoarseNet, and $w_{csd}$ and $w_{cph}$ are loss weights.
As a result, our \textit{inner loss} can be calculated by summing $\mathcal{L}_f$ and $\mathcal{L}_c$  with a weight $\lambda$ together:
\begin{equation}
\begin{aligned}
     & \mathcal{L}_{f+c}=\mathcal{L}_{f}+\lambda\mathcal{L}_{c}.
\end{aligned}
\label{eq:loss}
\end{equation}

For each optimization, we extract three consecutive frames in a video at a time and take the the first two frames as our meta-training set $\mathcal{D}_t$ and the last two frames as our meta-testing set $\mathcal{D}_{t+1}$.
Then we compute gradients from $\mathcal{L}_{f+c}$ in meta-training set and employ SGD to adapt our parameters of the model.
After each adaptation, we utilize adapted parameters to predict the depth and pose, then calculate gradients based on $\mathcal{L}_{f}$ in meta-testing set, and finally use the Adam optimizer~\cite{kingma2014adam} to optimize the original model.
This meta-learning algorithm incorporates the process of online adaptation into meta-training and ensures the high performance of the model in meta-testing. 
Alg.~\ref{alg:bml} shows the process of the basic meta-learning for a batch.
\IncMargin{1em}
\begin{algorithm}
    \algsetup{linenosize=\footnotesize}
    \footnotesize
    \SetAlgoNoLine 
    \SetKwInOut{Input}{\textbf{Require}}
 
\Input{\\
        Training set: $\mathcal{D}_s$;\\
        Hyper-parameters: $\alpha, \beta$\\
        Model parameters to be trained: $\theta$
        }
    \BlankLine
 
    initialize $\theta$\\
    \While{not done}{
        $\{{\mathcal{D}^b_{t}, \mathcal{D}^b_{t+1}}\} \sim \mathcal{D}_s$ \hfill $\rhd$Sample a batch\\
        ${\theta}_{t} \leftarrow \theta $  \hfill $\rhd$Initialize parameters of model\\
        $\mathcal{L}_{in} \leftarrow 0$  \hfill $\rhd$Initialize the inner loss\\
        $\mathcal{L}_{out} \leftarrow 0$ \hfill $\rhd$Initialize the outer loss\\
        \For {$i \leftarrow 1,...,b$}{
            $\mathcal{L}_{in} \leftarrow \mathcal{L}_{in} + \mathcal{L}_{f+c}({\theta_t, \mathcal{D}^i_{t}})$ \hfill $\rhd$ Accumulate $\mathcal{L}_{in}$\\
        }
        ${\theta}_{t+1} \leftarrow {\theta}_t -\alpha\nabla_{{\theta}_t}\mathcal{L}_{in}$ \hfill $\rhd$ Adaptation for $\theta_t$\\
        \For {$i \leftarrow 1,...,b$}{
            $\mathcal{L}_{out} \leftarrow \mathcal{L}_{out} + \mathcal{L}_f({\theta_{t+1}, \mathcal{D}^i_{t+1}})$ \hfill $\rhd$ Accumulate $\mathcal{L}_{out}$\\
        }
        $\theta \leftarrow \theta_t - \beta\nabla_{\theta_t} \mathcal{L}_{out}$ \hfill $\rhd$ Optimization for $\theta$\\
    }
    \caption{Basic meta-learning (BML) for depth completion.}
\label{alg:bml}
\end{algorithm}
\DecMargin{1em}

\begin{figure*}[!t]
    \centering
    \includegraphics[scale=0.45]{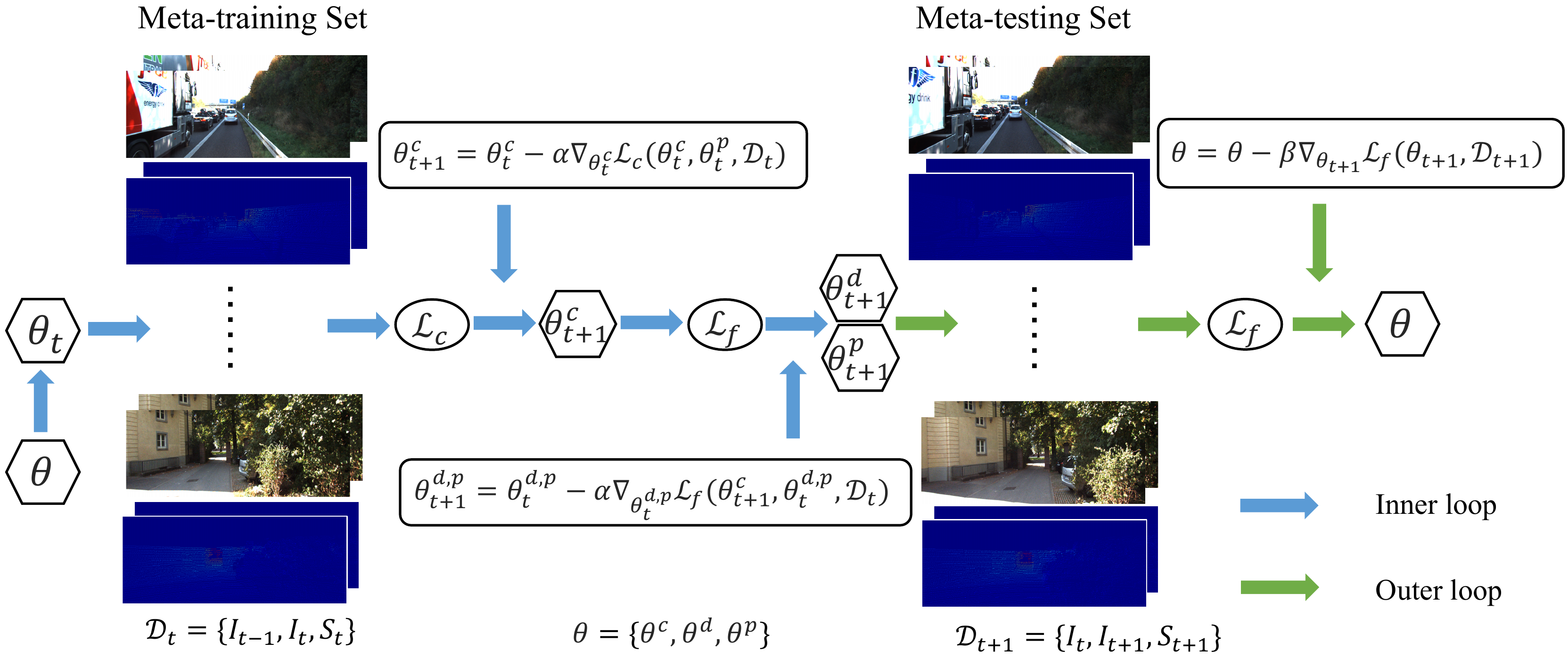}
    \caption{The framework of disentangled meta-learning(DML). \{$\theta^c, \theta^d, \theta^p\}$ are parameters of CoarseNet, DepthNet and PoseNet. $\theta$ is the original parameters of the model, while $\theta_t$ and $\theta_{t+1}$ refer to parameters used in inner-loop and outer-loop. In inner-loop, DML first adapts $\theta^c_t$ to $\theta^c_{t+1}$ by $\mathcal{L}_c$ and then uses $\theta^c_{t+1}$ to compute $\mathcal{L}_f$ to adapt $\theta^d_t$ and $\theta^p_t$, thus, $\theta_{t+1}$ is obtained. In outer-loop, $\mathcal{L}_f$ is utilized again to update $\theta$ based on $\theta_{t+1}$.}.
    \label{meta}
\label{fig:mlb}
\end{figure*}

\noindent\textbf{Disentangled meta-learning.}
Although the BML algorithm provides a solution into training a base model easy to fine-tune online,
it does not consider how to balance the influence of different modal information on the model.
Differences in the structure and form of the two modal data may cause the two modal inputs to change differently when exposed to new environments, \textit{e.g.} depth input varies in sparsity while image input varies in content. 
Therefore, we consider that it is necessary to adjust to the changes of two modal inputs separately.
Fig.~\ref{fig:mlb} elaborates the process of disentangled meta-learning (DML) for two modal inputs,
which  divides the meta-training stage into two steps by adapting the model to depth input and image input respectively.
In detail, we first adapt our CoarseNet to the changes of depth sparsity and then adapt our PoseNet and DepthNet to the changes of image content.
In the first adaptation to sparse input, we utilize the $\mathcal{L}_c$ in Eq.~\ref{eq:lossc} to update the CoarseNet using SGD on $\mathcal{D}_{t}$.
The aim of first adaptation step is to adapt the CoarseNet to changes in sparse depth input to produce a more precise depth map with full-density for DepthNet.
In the second adaptation, we use adapted CoarseNet to compute the $\mathcal{L}_f$ in Eq.~\ref{eq:lossf} and adapt our DepthNet and PoseNet using SGD on $\mathcal{D}_{t}$.
Since the CoarseNet has been adapted to changes in depth sparsity, the second adaptation can be more focused on changes in image content.
After two steps of meta-training, we apply the adapted model to predict depth on $\mathcal{D}_{t+1}$ and compute $\mathcal{L}_f$ in Eq.~\ref{eq:lossf}, and finally optimize our original model based on gradients from $\mathcal{L}_f$ in meta-testing.
We provide the pseudo-code of our disentangled meta-learning for a batch in Alg.~\ref{alg:dml}.
\IncMargin{1em}
\begin{algorithm}
   \algsetup{linenosize=\footnotesize}
    \footnotesize
    \SetAlgoNoLine 
    \SetKwInOut{Input}{\textbf{Require}}
 
\Input{\\
        Training set: $\mathcal{D}_s$;\\
        Hyper-parameters: $\alpha, \beta$\\
        Model parameters to be trained: $\theta=\{\theta^c, \theta^d, \theta^p\}$\\
        }
    \BlankLine
 
    initialize $\theta$\\
    \While{not done}{
        $\{{\mathcal{D}^b_{t}, \mathcal{D}^b_{t+1}}\} \sim \mathcal{D}_s$ \hfill $\rhd$Sample a batch\\
        $\{{\theta}^c_{t},{\theta}^d_{t},{\theta}^p_{t}\} \leftarrow \theta $  \hfill $\rhd$Initialize parameters of model\\
        $\mathcal{L}_{in1}, \mathcal{L}_{in2} \leftarrow 0$ \hfill $\rhd$Initialize two inner losses\\
        $\mathcal{L}_{out} \leftarrow 0$ \hfill $\rhd$Initialize the outer loss\\
        \For {$\mathcal{D}^b_{t}$}{
            $\mathcal{L}_{in1} \leftarrow \sum_{i=1}^b \mathcal{L}_c({\theta_t^c,\theta_t^p, \mathcal{D}^i_{t}})$ \hfill $\rhd$ Compute $\mathcal{L}_{in1}$\\
            ${\theta}_{t+1}^c \leftarrow {\theta}_t^c -\alpha\nabla_{{\theta}_t^c}\mathcal{L}_{in1}$ \hfill $\rhd$ Adapt $\theta_t^c$\\
            $\mathcal{L}_{in2} \leftarrow \sum_{i=1}^b \mathcal{L}_f({\theta_{t+1}^c,\theta_t^p, \theta_t^d, \mathcal{D}^i_{t}})$ \hfill $\rhd$ Compute $\mathcal{L}_{in2}$\\
            ${\theta}_{t+1}^d \leftarrow {\theta}_t^d -\alpha\nabla_{{\theta}_t^d}\mathcal{L}_{in2}$ \hfill $\rhd$ Adapt $\theta_t^d$\\
            ${\theta}_{t+1}^p \leftarrow {\theta}_t^p -\alpha\nabla_{{\theta}_t^p}\mathcal{L}_{in2}$ \hfill $\rhd$ Adapt $\theta_t^p$\\
        }
        $\theta_{t+1} \leftarrow \{{\theta}^c_{t+1},{\theta}^d_{t+1},{\theta}^p_{t+1}\}$\\
        \For {$\mathcal{D}^b_{t+1}$}{
            $\mathcal{L}_{outer} \leftarrow \sum_{i=1}^b \mathcal{L}_f({\theta_{t+1}, \mathcal{D}^i_{t+1}})$ \hfill $\rhd$ Compute  $\mathcal{L}_{out}$\\
        }
        $\theta \leftarrow \theta_t - \beta\nabla_{\theta_t} \mathcal{L}_{out}$ \hfill $\rhd$ Optimization for $\theta$\\
    }
    \caption{Disentangled Meta-learning (DML) for two modal inputs.}
\label{alg:dml}
\end{algorithm}
\DecMargin{1em}
\subsection{Online adaptation for depth completion}
\begin{figure*}
    \centering
    \includegraphics[scale=0.5]{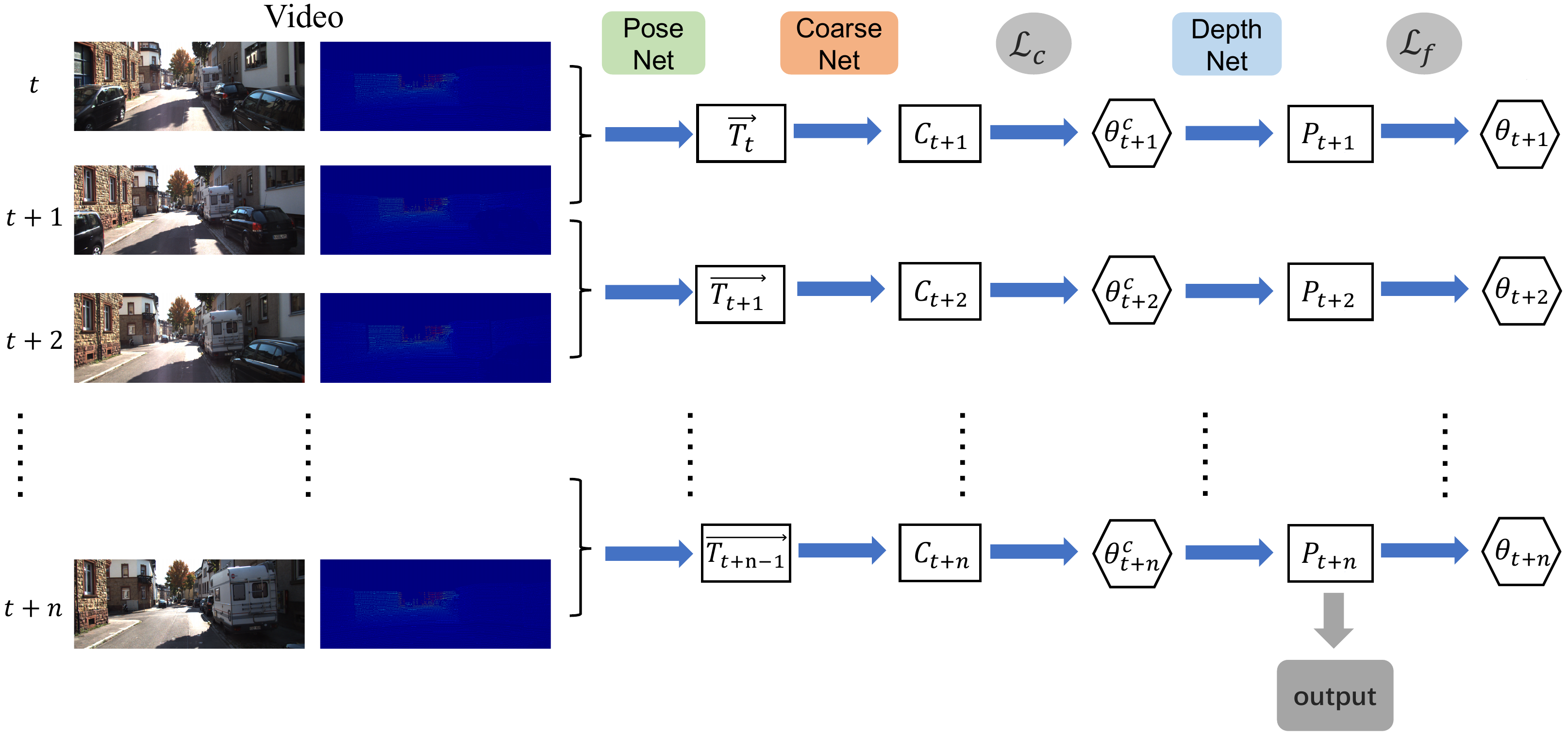}
    \caption{Online adaptation on a target video. $\protect\overrightarrow{T_t}$ indicates the relative pose from the $t+1$ frame to the $t$ frame. It implements real-time adaptation of $\theta$ of the model by first predicting $\protect\overrightarrow{T_t}$ from PoseNet for computation of $\mathcal{L}_c$ and $\mathcal{L}_f$, then updating $\theta_c$ of CoarseNet according to $\mathcal{L}_c$, and finally updating $\theta_d$ of DepthNet and $\theta_p$ of PoseNet according to $\mathcal{L}_f$ computed using updated $\theta_c$.}
\label{fig:oa}
\end{figure*}

We follow online adaptation in \cite{l2a}. In specific, for a video sequence, we test the performance for each new frame and then update the model's parameters according to the performance on the current frame.
Each optimization in the current frame is served for the test on next frame in the unseen video.
In online depth completion adaptation, self-supervised loss function is used for a gradient descent step.
Basic online adaptation step can written as:
\begin{equation}
    \theta_t \leftarrow \theta_{t-1} - \gamma \nabla_{\theta_{t-1}}\mathcal{L}(\theta_{t-1},\mathcal{D}_t),
\end{equation}
where $\gamma$ is the learning rate in online adaptation, $\theta_t$ is the parameters of model in the $t$ frame in a video and $\theta_0$ is the parameters of the base model.
Considering the inputs of depth completion are multi-modal, in light of  our DML algorithm, we modify this online adaptation paradigm as follows,
\begin{equation}
    \begin{aligned}
       & \theta^c_t \leftarrow \theta^c_{t-1} - \gamma \nabla_{\theta^c_{t-1}}\mathcal{L}_{c}(\theta^c_{t-1},\theta^p_{t-1},\mathcal{D}_t), \\
&\theta^{d}_t \leftarrow \theta^{d}_{t-1} - \gamma \nabla_{\theta^{d}_{t-1}}\mathcal{L}_{f}(\theta^c_{t},\theta^d_{t-1},\theta^p_{t-1},\mathcal{D}_t),\\
&\theta^{p}_t \leftarrow \theta^{p}_{t-1} - \gamma \nabla_{\theta^{p}_{t-1}}\mathcal{L}_{f}(\theta^c_{t},\theta^d_{t-1},\theta^p_{t-1},\mathcal{D}_t),\\
    \end{aligned}
\end{equation}
where $\theta^c, \theta^d, \theta^p$ are referred to parameters of CoarseNet, DepthNet and PoseNet respectively.
Compared to basic online adaptation, our online adaptation has two steps including updating CoarseNet first and then updating PoseNet and DepthNet.
The first step aims to adapt our model to changes of depth input so that stable depth input can be provided for step two.
After that, our model can be focused on the changes of image input in the second step.
This disentanglement allows the model to be more flexible in dealing with significant changes in one modality at one step, rather than treating the changes in two modal inputs as the same.
Fig.~\ref{fig:oa} illustrates our online adaptation procedure on a target video. Due to the computation of $\mathcal{L}_c$ and $\mathcal{L}_f$ both needing relative pose, we calculate relative pose through PoseNet before starting the two-step adaptation.

\section{Experiment}
To comprehensively analyze the effect of our method, we provide the experimental results about adaptation performance of models using different methods. In addition, we implement experiments to check the capacity of MetaComp to adapt to changes in different modal inputs, \textit{i.e.} adaptability to depth sparsity and adaptability to image content.
\subsection{Datasets}
\noindent\textbf{Virtual KITTI 2.} 
The VKITTI2 \cite{vkitti2} dataset updated from \cite{vkitti} is a synthetic dataset for urban driving environment.
It consists of five scenes with various conditions, \textit{e.g.,} sunset, overcast, rain, \textit{etc.,}, and has 85k images with dense depth maps. 
Since this paper focuses on the self-supervised depth completion, we thus generate the sparse depth maps from the available dense depth map by utilizing binomial distribution to randomly sample pixels from ground truth (5\%).
We employ it as our source domain and pre-train our model on it with our proposed method in the experiment for testing the adaption performance. Furthermore, since it can provide images of different conditions and generate sparse depth maps of different densities, it is also used in experiments about adapting to depth sparsity and image content.

\noindent\textbf{KITTI.} 
The KITTI \cite{kitti} dataset is a real-world dataset for car driving. The KITTI depth completion benchmark~\cite{sparse1} contains 93k raw images, sparse depth maps and semi-dense ground truth. 
The sparse depth maps are captured by Velodyne Lidar sensor with a 5\% density. 
The benchmark is split into 86k images for training, 7k images for validation \cite{sparse1}.
We take the KITTI validation set as our target domain including 26 videos with 7k frames in total and online adapt our model on it.
Besides, the KITTI training set is utilized in the experiment to adapt to depth sparsity.


\noindent\textbf{VOID.} 
VOID dataset is a real-world dataset for common life scenes, which consist of mostly indoor scenes (laboratories, classrooms) and a small part of outdoor scenes (gardens) \cite{void}.
It contains 56 video sequences providing a total of 49k frames of RGB images, sparse depth maps and dense depth maps, and is divided into the training set of 48 sequences and testing set of 8 sequences.
Each sparse depth map has about 1500 sparse points which covers 0.5\% of the image.
We include the VOID training set into our source domain to validate the performance of our method when the target domain is significantly different from the source domain.

\subsection{Implementation details}
We implement our framework in PyTorch. 
We build a baseline by training the model in a standard self-supervised learning framework($\mathcal{L}_f$) without meta-learning used.
We adopt Adam optimizer with an initial learning rate set to $1e-4$ to train our model for 15 epochs for training the baseline model. For training with meta-learning, we set our initial learning rate to $\alpha=1e-7$ and $\beta=1e-4$ using SGD and Adam in meta-training and meta-testing for 10 epochs, respectively. In online adaptation, we select learning rate as $\gamma=1e-6$ with Adam optimizer to update our networks. In addition, we choose $w_{sd}=1$, $w_{ph}=2$ and $w_{sm}=0.1$ for $\mathcal{L}_f$, set $w_{csd}=1$ and $w_{cph}=2$ for $\mathcal{L}_c$, and select $\lambda = 0.1$ for $\mathcal{L}_{f+c}$.

\begin{figure*}[!t]
    \centering
    \includegraphics[width=18 cm,height=5.5 cm]{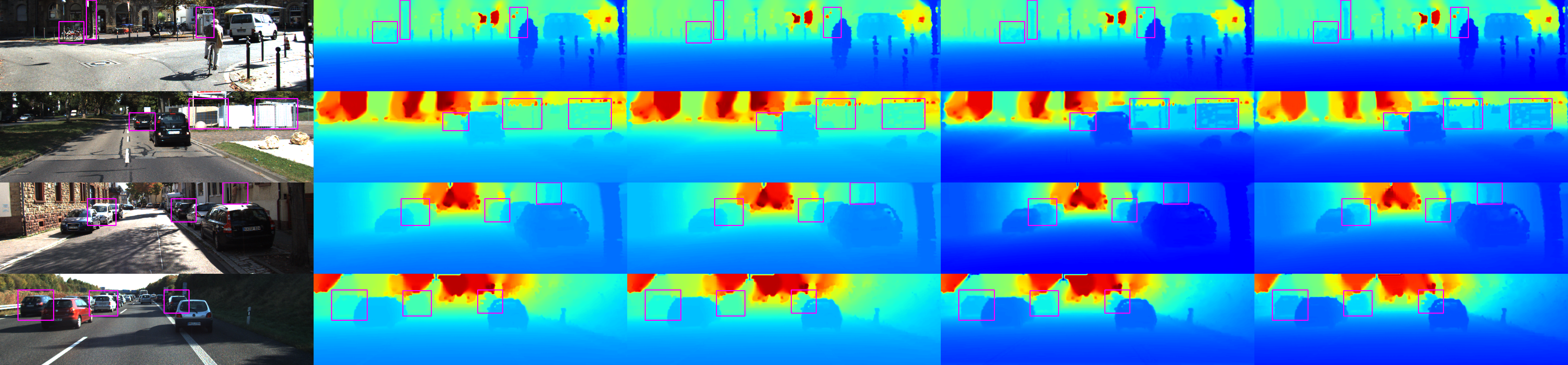}
    \begin{flushleft}
    	\small \qquad\qquad\quad Image\qquad\qquad\qquad\qquad\quad $\mathcal{L}_f$ \qquad\qquad\qquad\qquad\quad $\mathcal{L}_f$+OA \qquad\qquad\qquad\quad BML+OA \qquad\qquad\qquad\quad DML+OA
    \end{flushleft}  
    \caption{Qualitative comparison of our methods. It provides RGB images and visualizes depth maps predicted by models using different methods. Best viewed in color.}
    \label{fig:visualization}
\end{figure*}

\begin{figure*}
\centering
\subfloat[\small $\mathcal{L}_f$ with OA]
{\includegraphics[width=4.55cm,height=3.9cm]{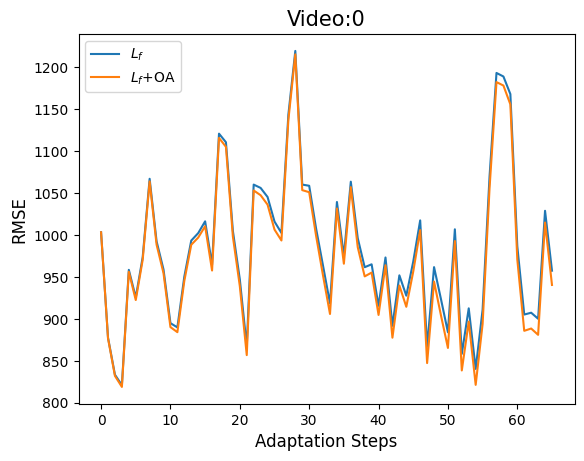}} 
\subfloat[\small DML with OA]
{\includegraphics[width=4.55cm,height=3.9cm]{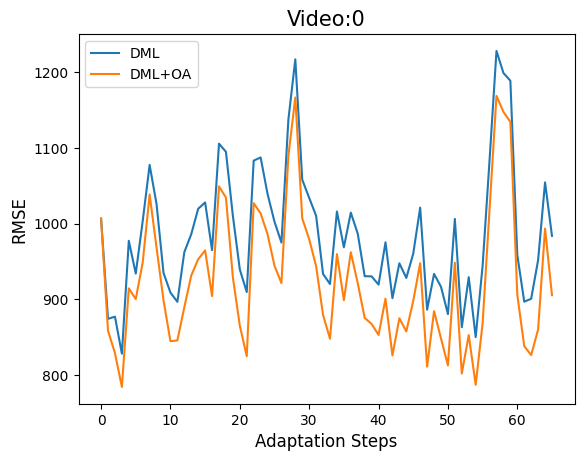}}
\subfloat[\small $\mathcal{L}_f$ and DML]
{\includegraphics[width=4.55cm,height=3.9cm]{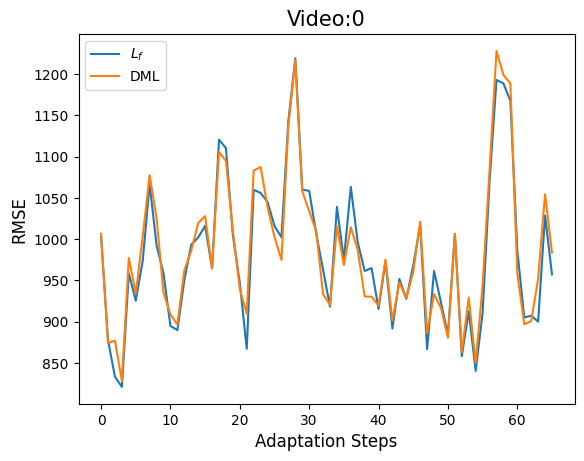}}
\subfloat[\small $\mathcal{L}_f$ and DML with OA]
{\includegraphics[width=4.55cm,height=3.9cm]{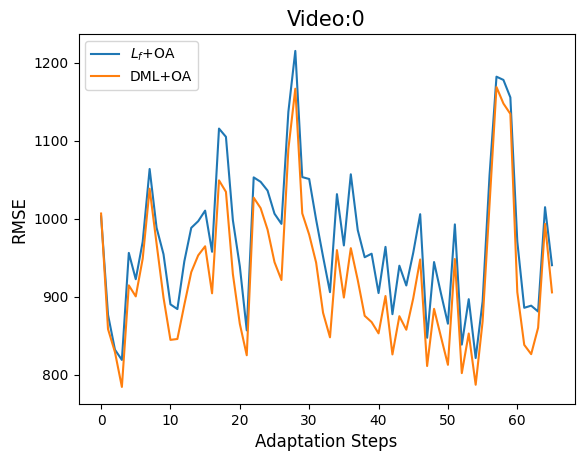}}
\caption{RMSE(mm) \textit{w.r.t.} the number of adaptation steps on Video 0 of 66 frames from KITTI validation set.
Figures show the performance of models on short video sequence and provide a comparison of $\mathcal{L}_f$ and DML with or without online adaptation (OA).
}
\label{fig:video0}
\end{figure*}

\subsection{Evaluation tool} 
To assess the performance of our framework, we utilize an evaluation tool appropriate for online adaptation.
We test the performance of our model in the current frame and then adapt our model before moving to the next frame in a video.
For the purpose of evaluating the performance of our model on different video sequences, we reset our model to the pre-trained one at the beginning of each video.
We calculate the average score on all the frames as our evaluation indicator for a video and take the average score of all videos as our final result.
We take MAE, RMSE, iMAE and iRMSE as our error metrics~\cite{sparse1}.

\subsection{Adaptation performance}
For evaluation, we pre-train our model on the source dataset, \textit{i.e.,} VKITTI, or VOID, and then test it on the target dataset, \textit{i.e.,} KITTI. 
We compare the performance of our model trained using self-supervised loss $\mathcal{L}_f$ (baseline), basic meta-learning framework (BML), and our disentangled meta-learning framework (DML), respectively.
For training methods with meta-learning, online adaptation adopts the same scheme of meta-training in inner loop. In detail, BML utilizes $\mathcal{L}_{f+c}$ to optimize our model online, while $\mathcal{L}_c$ and $\mathcal{L}_f$ are used successively to update parameters of CoarseNet and parameters of DepthNet and PoseNet for DML in online adaptation.
For model trained using self-supervised loss, $\mathcal{L}_f$ is still employed to adapt our network.

\begin{table}[htbp]\scriptsize
  \centering
  \caption{Quantitative results on KITTI pre-trained on VKITTI}
\begin{threeparttable}
\renewcommand{\arraystretch}{1.2}
\setlength{\tabcolsep}{4mm}{
\begin{tabular}{l||c|c|c|c}
\hline
Method & RMSE & MAE & iRMSE & iMAE \\
\hline
\hline
(a) $\mathcal{L}_f$(-) & 3541.56 & 1047.34 & 13.72 & 5.59 \\
(b) $\mathcal{L}_f$(-)+OA & 3174.24 & 938.06 & 11.29 & 5.21 \\
\hline
(c) BML(-) & 3501.03 & 875.27 & 14.51 & 6.55 \\
(d) BML(-)+OA & 3002.76 & 834.48 & 10.86 & 5.31\\
\hline
\hline
(e) $\mathcal{L}_f$ & 1411.05 & 474.63 & 5.07 & 2.82 \\
(f) $\mathcal{L}_f$+OA & 1401.89 & 385.31 & 4.37 & 1.93 \\
\hline
(g) BML& 1410.75 & 440.45 & 4.50 & 2.60 \\
(h) BML+OA & 1388.43 & 354.23 & \textbf{4.16} & 1.83 \\
\hline
(i) DML & 1378.46 & 505.30 & 5.23 & 3.02\\
(j) DML+OA & \textbf{1346.64} & \textbf{336.95} & 4.47 & \textbf{1.72}\\
\hline
\end{tabular}}
\begin{tablenotes}
	\footnotesize
    \item Notation (-) means that CoarseNet is not in the model. 
    The results illustrate the effectiveness of our framework on online adaptation (OA).
\end{tablenotes}
\end{threeparttable} 
\label{tab:adaptation}
\end{table}

Table.~\ref{tab:adaptation} shows the quantitative results of adaptation performance of different frameworks pre-trained on VKITTI. To show the advantage of CoarseNet on online adaptation, we additionally provide the testing results of the model only including DepthNet and PoseNet in (a)-(d).
It is found that our method, \textit{i.e.,} DML+OA, achieves the best performance on RMSE, MAE and iMAE, which proves the effectiveness of our framework on online adaptation. 
The comparisons between the methods with / without online adaptation (OA) indicate that the online adaptation can bring performance improvements no matter whether the meta-learning strategy is used. 
However, by comparing (a)-(b) with (c)-(d) and (e)-(f) with (g)-(h), we can find the meta-learning technique can yield more improvements. 
Through exploiting the CoarseNet to convert a sparse depth to a coarse dense map, we can observe remarkable performance improvements by comparing (a)-(d) with (e)-(h), which demonstrates that the deployment of CoarseNet can boost the performance significantly. 
Lastly, two important comparisons, \textit{i.e.,}  (e)-(f) \textit{v.s.} (i)-(j) and (g)-(h) \textit{v.s.} (i)-(j), shows the superiority of our DML. Fig.~\ref{fig:visualization} shows the qualitative comparison of different methods through visualizing predicted depth maps.

\begin{figure*}[!t]
\centering
\includegraphics[width=15cm,height=5.5cm]{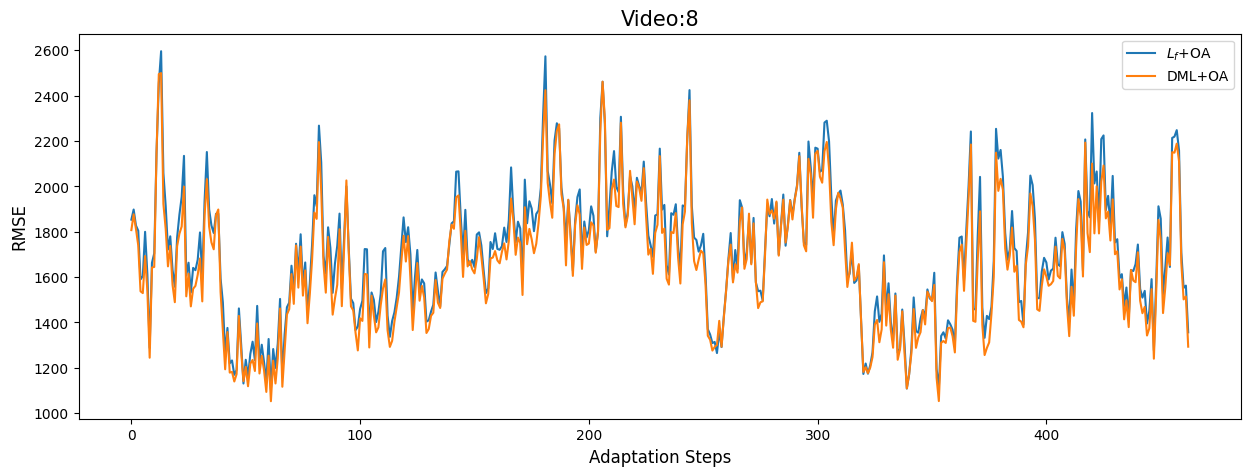}
\caption{RMSE(mm) \textit{w.r.t.} the number of adaptation steps on Video 8 of 463 frames from KITTI validation set. 
The figure shows the performance of models on long video sequence and provide a comparison of $\mathcal{L}_f$ and DML with online adaptation (OA).
}
\label{fig:video8}
\end{figure*}

To further analyse the effect of our method on videos with different lengths, we show the process of online adaptation in Fig.~\ref{fig:video0} and Fig.~\ref{fig:video8}. Fig.~\ref{fig:video0}(a) and (b) illustrate the notable improvement brought by DML for online adaptation. Adapting the model directly trained in a self-supervised manner sometimes only brings a slight improvement, while DML can ameliorate the adaptation and then improve the model significantly during online adaptation.
Fig.~\ref{fig:video0}(d) provides a more clear comparison for the two methods.
Besides, without online adaptation, DML can still achieve an initial performance that is not inferior to $\mathcal{L}_f$, which can be seen in Fig.~\ref{fig:video0}(c).
Fig.~\ref{fig:video8}. reveals our method's capability of adapting to long video sequences. 

\begin{table}[htbp]\scriptsize
  \centering
  \caption{Quantitative results on KITTI pre-trained on VOID.}
\renewcommand{\arraystretch}{1.2}
\setlength{\tabcolsep}{4mm}{
\begin{tabular}{c||c|c|c|c}
\hline
Method & RMSE & MAE & iRMSE & iMAE \\
\hline
\hline
$\mathcal{L}_f$ & 9983.87 &	5659.48	& 174.20 & 16.61 \\
$\mathcal{L}_f$+OA & 2889.20 & 1840.36 & 138.42	& 15.78\\
\hline
BML& 3805.99 & 2438.95 & 11.09 & 8.49\\
BML+OA & 1529.26 & 576.17 & 5.50 & 2.77\\
\hline
DML & 1453.06 & 474.67 & 4.90 & 2.47\\
DML+OA & \textbf{1376.50} & \textbf{416.07} & \textbf{4.86} & \textbf{1.93}\\
\hline
\end{tabular}}
\label{tab:void}
\end{table}

The above experiment demonstrates the effectiveness of our method when the source and target domains are both car driving datasets with similar scenes.
To prove our method is applicable when the source and target domains are distinct, we pre-train our model on VOID training set and online adapt it on KITTI validation set, the results are shown in Table.~\ref{tab:void}. 
VOID dataset and KITTI dataset differ greatly in the range of depth values, density of depth points, image resolution and scene content.
Therefore, it is found that the initial performance of the model trained using $\mathcal{L}_f$ is very poor. 
In spite of a significant improvement brought by online adaptation, $\mathcal{L}_f$+OA still does not meet our expectation. 
BML alleviates this situation, and BML+OA achieves passable performance.
While DML further enhances the capacity of the model to adapt to the gap between source and target domain and DML+OA achieves the best performance which is close to the best one in Table~\ref{tab:adaptation} showing the advantage of DML for model adaptation to new environments.

\subsection{Adaptability to depth sparsity}
To validate the capability of fast adaptation of our model to depth sparsity, we modify the density of sparse maps in VKITTI to re-train our model and then sample sparse points from sparse maps of KITTI validation set with different ratios to form new sparse inputs for testing.
Tab.~\ref{tab:sparsity} shows the evaluation results on different densities for validating the adaptability of our model to depth sparsity.

\begin{table}[htbp]\scriptsize
  \centering
  \caption{RMSE on KITTI for sensitivity analysis of sparisy.}
\begin{threeparttable}
\renewcommand{\arraystretch}{1.2}
\setlength{\tabcolsep}{4mm}{
\begin{tabular}{c|c||c|c|c}
\hline
Training & \multirow{2}*{Method} & \multicolumn{3}{c}{Testing Density}\\
\cline{3-5}
{Density} & {} & 1\% & 2.5\% & 5\% \\
\hline
\hline
\multirow{4}*{1\%} & $L_f$ & 1912.66 & 1572.61 & 1418.42 \\
{} & $\mathcal{L}_f$+OA & 1919.99* & 1570.21 & 1404.78 \\
{} & BML+OA & 1883.60 & 1549.55 & 1397.79 \\
{} & DML+OA & 1863.69 & 1538.80 & 1393.47 \\
\hline
\hline
\multirow{4}*{2.5\%} & $L_f$ & 1951.59 & 1602.79 & 1450.70 \\
{} & $\mathcal{L}_f$+OA & 1916.43 & 1536.46 & 1387.48 \\
{} & BML+OA & 1893.19 & 1523.99 & 1366.94 \\
{} & DML+OA & 1825.23 & 1485.14 & 1343.39 \\
\hline
\hline
\multirow{4}*{5\%} & $L_f$ & 1914.29 & 1571.83 & 1411.05 \\
{} & $\mathcal{L}_f$+OA & 1907.37 & 1565.78 & 1401.89 \\
{} & BML+OA & 1893.65 & 1545.98 & 1386.23 \\
{} & DML+OA & 1879.72 & 1514.24 & 1346.64 \\
\hline
\hline
\multirow{4}*{10\%} & $L_f$ & 2100.79 & 1663.79 & 1451.45\\
{} & $L_f$+OA & 2096.97 & 1657.87 & 1440.37 \\
{} & BML+OA & 2010.10 & 1613.99 & 1420.03 \\
{} & DML+OA & 1954.44 & 1568.21 & 1397.40 \\
\hline
\end{tabular}}
\begin{tablenotes}
	\footnotesize
	\item We train our model on VKITTI and then test it on KITTI with different training densities and testing densities. Notation(*) means taht online adaptation brings no improvement.
\end{tablenotes}
\end{threeparttable}
\label{tab:sparsity}
\end{table}

\begin{table}[htbp]\scriptsize
  \centering
  \caption{RMSE on KITTI without CoarseNet for sensitivity analysis of sparsity.}
\begin{threeparttable}
\renewcommand{\arraystretch}{1.2}
\setlength{\tabcolsep}{3.5mm}{
\begin{tabular}{c||c|c|c|c}
\hline
\multirow{2}*{Method} & \multicolumn{4}{c}{Training Density}\\
\cline{2-5}
 {} & 1\% & 2.5\% & 5\% & 10\%\\
\hline
$\mathcal{L}_f$(-) & 12174.42 & 7476.42 & 3541.56 & 2994.40 \\
$\mathcal{L}_f$(-)+OA & 12242.63* & 6535.89 & 3174.24 & 2866.24 \\
\hline
BML(-) & 13887.81 & 12406.19 & 3501.03 & 3217.22 \\
BML(-)+OA & 14134.00* & 12427.32* & 3002.76 & 2714.95 \\
\hline
\end{tabular}}
\begin{tablenotes}
	\footnotesize
	\item We train the model without CoarseNet on VKITTI and then test it on KITTI. Training density varies from 1\% to 10\%, while the testing density is fixed to 5\%. Notation (-) means that CoarseNet is not in the model. Notation(*) means that online adaptation brings no improvement.
\end{tablenotes}
\end{threeparttable}
\label{tab:sparsity_noc}
\end{table}

\begin{table}[htbp]\scriptsize
  \centering
  \caption{RMSE on videos of Scene01 of VKITTI.}
\begin{threeparttable}
\renewcommand{\arraystretch}{1.2}
\setlength{\tabcolsep}{0.5mm}{
\begin{tabular}{c||c|c|c|c|c|c|c|c}
\hline
\multirow{2}*{Method} & \multicolumn{6}{c|}{Condition} & \multirow{2}*{Mean} & Standard \\
\cline{2-7}
{} & Clone & Fog & Morning & Overcast & Rain & Sunset & {} & Deviation\\
\hline
$L_f$ & 5697.57 & 5702.62 & 5695.82 & 5687.62 &	5699.44 & 5699.83 &	5697.15 & 5.20\\
\hline
$L_f$+OA  & 5512.82 & 5468.12 & 5491.30 & 5491.30 &	5471.56 & 5509.25 &	5490.73 & 18.49\\
BML+OA & 5438.80 & \textbf{5388.72} & 5443.04 & 5421.45 & \textbf{5392.77} & 5460.09 & 5424.14 & 28.67\\
DML+OA & \textbf{5418.51} & 5399.64 & \textbf{5415.77} & \textbf{5408.48} & 5403.31 & \textbf{5415.45} & \textbf{5410.19} & \textbf{7.61}\\
\hline
\end{tabular}}
\begin{tablenotes}
	\footnotesize
	\item We train our model on KITTI training set and then test it on videos of Scene01 of VKITTI with different conditions.
\end{tablenotes}
\end{threeparttable}
\label{tab:appearance}
\end{table}

From Tab.~\ref{tab:sparsity}, we can find that no matter how the training density and testing density change, our framework, \textit{i.e.,} DML+OA, always achieves the best performance, which proves the robustness of our method to sparsity.
For the extreme case of 1\% density in both training and testing, online adaptation dose not work when the model is trained by self-supervised loss, while DML with OA still significantly improves the performance, which reveals the capability of our model adapting to low density.
Moreover, when the density is greatly reduced from training to testing, \textit{e.g.}, from 10\% in training to 1\% in testing, all methods suffer a serious performance drop, while ours DML+OA still performs better.

In contrast, for the model without CoarseNet, meta-learning only works when training density is high as shown in Tab.~\ref{tab:sparsity_noc}. When the training density is 5\% or 10\%, online adaptation is effective for models trained by either self-supervised loss or BML. 
Furthermore the error metric decreases more when BML is utilized. 
However, when training density is in a low level, \textit{e.g.} 2.5\%, BML makes the model not converge because of its complex training process.
When training density decreases to 1\%, two methods both fail on online adaptation.
It shows the the sensitivity of the model without CoarseNet to sparsity and the difficulty for the model to adapt to changes of multi-modal inputs when sparsity varies greatly.
Therefore, we introduce CoarseNet and disentangle the online adaptation to first adapt CoarseNet to depth sparsity variation to ensure the whole model can adapt to changes of multi-modal inputs.


\subsection{Adaptability to image content}

To test the ability of our model to adjust to images content, we pre-train our model on KITTI training dataset and then online adapt our model on VKITTI with both a density of 5\%.
The VKITTI dataset is rendered from KITTI videos by directly cloning or modifying conditions such as weather, lighting, etc.
We select Scene01 in VKITTI as our target domain. Scene01 contains several video sequences providing the same depth maps and different images in various conditions, such as clone, fog, morning, overcast, rain, and sunset.
Results are shown in Tab.~\ref{tab:appearance}. Our DML framework achieves the best performance in most of conditions. It demonstrates the superiority of our DML in adapting to content changes, which also can be proved by the mean of RMSE.
In addition, DML framework also achieves the lowest standard deviation of RMSE when online adaptation is used, showing the stability of our DML framework to different conditions.

\section{Conclusion}
In this paper, we studied the problem of online adaptation for depth completion with two-modal inputs through examining the meta-learning technique. To adapt the pre-trained depth completion model to a new environment with online multi-modal data, we disentangle the meta-training stage in basic meta-learning algorithm into two steps, one focusing on the sparse depth data while the other one aiming at adapting the model to new image content. In this way, our proposed DML can enable the pre-trained model to adjust its parameters effectively in a new environment. The experimental results and comprehensive analysis show the effectiveness of the basic and the developed disentangled meta-learning algorithm in online depth completion adaptation while our DML performs better and is more robust to the changes in multi-modal inputs.
\balance

\end{document}